%% file: main.tex
\documentclass[11pt,a4paper]{article}
\usepackage[hyperref]{eacl2021}
\usepackage{times}
\usepackage{latexsym}
\renewcommand{\UrlFont}{\ttfamily\small}

\usepackage{microtype}

\aclfinalcopy 
\def\aclpaperid{486} 

\usepackage{url}
\usepackage{amsmath,amsfonts,amssymb}
\usepackage{xspace}
\usepackage{subcaption}
\usepackage{appendix}
\usepackage{etoolbox}
\usepackage{stfloats}
\BeforeBeginEnvironment{appendices}{\clearpage}
\usepackage{microtype}
\usepackage{graphicx}
\usepackage[ruled,vlined]{algorithm2e}
\usepackage{enumitem}
\usepackage{algpseudocode}
\usepackage{floatrow}
\newfloatcommand{capbtabbox}{table}[][\FBwidth]

\usepackage{blindtext}

\usepackage{booktabs}
\usepackage{xspace}
\usepackage{placeins}
\newcommand{\Atis}{\mbox{\textsc{Atis}}\xspace}
\newcommand{\Geo}{\mbox{\textsc{GeoQuery}}\xspace}
\newcommand{\Jobs}{\mbox{\textsc{Jobs}}\xspace}

\newcommand{\gen}[1]{\textit{GEN} [#1]\xspace}
\newcommand{\system}[1]{\textsc{#1}\xspace}
\newcommand{\reduce}[1]{\system{REDUCE} [#1]\xspace}
\newcommand{\parser}{\texttt{ProtoParser}\xspace}
\newcommand{\pre}[1]{\textsl{#1}\xspace}
\newcommand{\data}[1]{\textsc{#1}\xspace}
\newcommand{\atis}{\data{ATIS}}
\newcommand{\geoquery}{\data{GeoQuery}}
\newcommand{\jobs}{\data{Jobs}}
\newcommand{\spider}{\data{Spider}}

\newcommand{\seqseq}{\system{Seq2Seq}}

\newcommand{\coarsefine}{\system{Coarse2Fine}}
\newcommand{\ptmaml}{\system{PT-MAML}}
\newcommand{\da}{\system{DA}}
\newcommand{\irnet}{\system{IRNet}}
\newcommand{\ratsql}{\system{RATSQL}}
\newcommand{\pt}{\textit{pt}}
\newcommand{\os}{\textit{os}}
\newcommand{\comb}{\textit{cb}}

\input{math_comm.tex}

\newcommand{\todo}[1]{\textcolor{red}{TODO}}

\title{Few-Shot Semantic Parsing for New Predicates}
\author{Zhuang Li, Lizhen Qu\thanks{corresponding author} ,  Shuo Huang, Gholamreza Haffari\\
Faculty of Information Technology\\
  Monash University \\
  {\tt firstname.lastname@monash.edu}\\
  {\tt shua0043@student.monash.edu}}

\date{}

\begin{document}

\maketitle

\begin{abstract}
\input{sec0_abs.tex} 
\end{abstract}

\section{Introduction}
\label{sec:intro}
\input{sec1_intro.tex}

\section{Related Work}
\input{sec2_related.tex}

\input{sec3_fewshot.tex}

\section{Experiments}
\input{sec4_expr.tex}

\section{Conclusion and Future Work}
\input{sec5_conc.tex}

\bibliography{anthology,emnlp2020,tacl2018}
\bibliographystyle{acl_natbib}

\appendix

\input{appendix}

\end{document}

%% file: math_comm.tex
\usepackage{amsthm,amsmath,amsfonts,bm,xspace}
\usepackage{color}

\newcommand{\comment}[1]{}

\def\eqref#1{(\ref{#1})}

\def\1{\bm{1}}

\def\va{{\bm{a}}}

\def\vx{{\bm{x}}}
\def\vy{{\bm{y}}}

\DeclareMathAlphabet{\mathsfit}{\encodingdefault}{\sfdefault}{m}{sl}
\SetMathAlphabet{\mathsfit}{bold}{\encodingdefault}{\sfdefault}{bx}{n}



%% file: sec0_abs.tex
In this work, we investigate the problems of semantic parsing in a few-shot learning setting. In this setting, we are provided with $k$ utterance-logical form pairs per new predicate. The state-of-the-art neural semantic parsers achieve less than 25\% accuracy on benchmark datasets when $k=1$. To tackle this problem, we proposed to i) apply a designated meta-learning method to train the model; ii) regularize attention scores with alignment statistics; iii) apply a smoothing technique in pre-training. As a result, our method consistently outperforms all the baselines in both one and two-shot settings.

%% file: sec1_intro.tex
%
%
Semantic parsing is the task of mapping natural language (NL) utterances to structured meaning representations, such as logical forms (LF). One key obstacle preventing the wide application of semantic parsing is the lack of task-specific training data. New tasks often require new predicates of LFs. Suppose a personal assistant (e.g. Alexa) is capable of booking flights. Due to new business requirement it needs to book ground transport as well. A user could ask the assistant "\textit{How much does it cost to go from Atlanta downtown to airport?}". The corresponding LF is as follows:

{\small
\begin{center}
    \begin{tabular}{c r}
    \label{example_template}
    \emph{(lambda \$0 e (exists \$1 (and ( ground\_transport \$1 )  }\\ 
     \emph{(to\_city \$1 atlanta:ci )(from\_airport \$1 atlanta:ci)}\\
     \emph{( =(ground\_fare \$1 ) \$0 ))))} 
    \end{tabular}
\end{center}
}
\noindent where both \textit{ground\_transport} and \textit{ground\_fare} are new predicates while the other predicates are used in flight booking, such as \textit{to\_city}, \textit{from\_airport}.
As manual construction of large parallel training data is expensive and time-consuming, we consider the \textit{few-shot} formulation of the problem, which requires only a handful of utterance-LF training pairs for each new predicate. The cost of preparing few-shot training examples is low, thus the corresponding techniques permit significantly faster prototyping and development than supervised approaches for business expansions.

Semantic parsing in the few-shot setting is challenging. In our experiments, the accuracy of the state-of-the-art (SOTA) semantic parsers drops to less than \textit{25\%}, when there is only one example per new predicate in training data.
Moreover, the SOTA parsers achieve less than 32\% of accuracy on five widely used corpora, when the LFs in the test sets do not share LF \emph{templates} in the training sets~\cite{finegan2018improvingtexttosql}. An LF template is derived by \emph{normalizing} the entities and attribute values of an LF into typed variable names~\cite{finegan2018improvingtexttosql}.
The few-shot setting imposes two major challenges for SOTA neural semantic parsers.  
First, it lacks sufficient data to learn effective representations for new predicates in a supervised manner. 
%
%
Second, new predicates bring in new LF templates, which are mixtures of known and new predicates. In contrast, the tasks (e.g. image classification) studied by the prior work on few-shot learning~\cite{snell2017prototypical,finn2017maml} considers an instance exclusively belonging to either a known class or a new class. Thus, it is non-trivial to apply conventional few-shot learning algorithms to generate LFs with mixed types of predicates.

To address above challenges, we present \parser, a transition-based neural semantic parser, which applies a sequence of parse actions to transduce an utterance into an LF template and fills the corresponding slots. The parser is pre-trained on a training set with known predicates, followed by fine-tuning on a \textit{support set} that contains few-shot examples of new predicates. It extends the attention-based sequence-to-sequence architecture~\cite{sutskever2014sequence} with the following novel techniques to alleviate the specific problems in the few-shot setting:
\begin{itemize}
    \item \textit{Predicate-droput}. Predicate-droput is a meta-learning technique to improve representation learning for both known and new predicates. We empirically found that known predicates are better represented with supervisely learned embeddings, while new predicates are better initialized by a metric-based few-shot learning algorithm~\cite{snell2017prototypical}. In order to let the two types of embeddings work together in a single model, we devised a training procedure called \textit{predicate-dropout} to simulate the testing scenario in pre-training.
    \item \textit{Attention regularization}. In this work, new predicates appear approximately once or twice during training. Thus, it is insufficient to learn reliable attention scores in the Seq2Seq architecture for those predicates. In the spirit of supervised attention~\cite{liu2016nmtSupervisedAttention}, we propose to regularize them with alignment scores estimated by using co-occurrence statistics and string similarity between words and predicates. The prior work on supervised attention is not applicable, because it requires either large parallel data~\cite{liu2016nmtSupervisedAttention}, significant manual effort~\cite{bao2018supervisedAttentionRationales,rabinovich2017semanticParsingSupervisedAttention}, or it is designed only for applications other than semantic parsing~\cite{liu2017eventDetectionSupervisedAttention,kamigaito2017supervisedAttentionConstituencyParsing}.
    \item \textit{Pre-training smoothing}. The vocabulary of predicates in fine-tuning is higher than that in pre-training, which leads to a distribution discrepancy between the two training stages. Inspired by Laplace smoothing~\cite{manning2008introductionIR}, we achieve significant performance gain by applying a smoothing technique during pre-training to alleviate the discrepancy.
\end{itemize}
Our extensive experiments on three benchmark corpora show that \parser outperforms the competitive baselines with a significant margin. The ablation study demonstrates the effectiveness of each individual proposed technique. The results are statistically significant with p$\le$0.05 according to the Wilcoxon signed-rank test~\cite{wilcoxon1992individual}.

%% file: sec2_related.tex
\paragraph{Semantic parsing}
There is ample of work on machine learning models for semantic parsing. The recent surveys~\cite{Kamath2018semanticParsingSurvey, zhu2019survey} cover a wide range of work in this area. The semantic formalism of meaning representations range from lambda calculas~\cite{montague1973lambdaCalculas}, SQL, to abstract meaning representation~\cite{banarescu2013AMR}. At the core of most recent models~\cite{chen2018seq2act, cheng2019executableParser, lin2019grammarText2SQL, zhang2019broadTransduction,yin2018tranx} is \seqseq with attention~\cite{bahdanau2014attention} by formulating the task as a machine translation problem. \coarsefine~\cite{dong2018coarse2fine} reports the highest accuracy on \geoquery~\cite{zelle1996geoQuery} and \Atis~\cite{price1990atis} in a supervised setting. \irnet~\cite{guo2019irnet} and \ratsql~\cite{wang2019rat} are two best performing models on the Text-to-SQL benchmark, \spider~\cite{yu2018spider}. They are also designed to be able to generalize to unseen database schemas.
However, supervised models perform well only when there is sufficient training data.
\paragraph{Data Sparsity} 
Most semantic parsing datasets are small in size. To address this issue, one line of research is to augment existing datasets with automatically generated data~\cite{su2017crossdomainsemanticparsing, jia2016datarecombination, cai2013openDomainFreebase}. Another line of research is to exploit available resources, such as knowledge bases~\cite{krishnamurthy2017typeConstraints,herzig2018decouplingZeroShot, chang2019zeroShotAuxiliaryTask,lee2019clauseCrossDomain,zhang2019editingSQLCrossDomain,guo2019irnet,wang2019rat}, semantic features in different domains~\cite{dadashkarimi2018zeroShotTransferSemanticParsing,li2020domain}, or unlabeled data~\cite{yin18aclvae, kovcisky2016semisupervised,sun2019neural}. Those works are orthogonal to our setting because our approach aims to efficiently exploit a handful of labeled data of new predicates, which are not limited to the ones in knowledge bases. Our setting also does not require involvement of humans in the loop such as active learning~\cite{duong2018activeLearning,ni2019mergingWeakActive} and crowd-sourcing~\cite{wang2015overnight, herzig2019dontParaphraseDetect}. We assume availability of resources different than the prior work and focus on the problems caused by new predicates. We  develop an approach to generalize to unseen LF templates consisting of both known and new predicates.

\paragraph{Few-Shot Learning} Few-shot learning is a type of machine learning problems that provides a handful of labeled training examples for a specific task. The survey~\cite{zhu2019survey} gives a comprehensive overview of the data, models, and algorithms proposed for this type of problems. It categorizes the models into multitask learning~\cite{hu2018multitaskLegalAttr}, embedding learning~\cite{snell2017prototypical,vinyals2016matchingnetworks}, learning with external memory~\cite{lee2018ExternalMemGradient,sukhbaatar2015endMemoryNetworks}, and generative modeling~\cite{reed2017generative} in terms of what prior knowledge is used. 
\cite{lee2019oneshot} tackles the problem of poor generalization across SQL templates for SQL query generation in the one-shot learning setting. In their setting, they assume all the SQL templates on test set are shared with the templates on support set. In contrast, we assume only the sharing of new predicates between a support set and a test set. In our one-shot setting, only around 10\% of LF templates on test set are shared with the ones in the support set of \geoquery dataset.

%% file: sec3_fewshot.tex
\section{Semantic Parser}
\label{sec:template-general}

\parser follows the SOTA neural semantic parsers~\cite{dong2018coarse2fine,guo2019irnet} to map an utterance into an LF in two steps: \textit{template generation} and \textit{slot filling}\footnote{Code and datasets can be found in this repository: \url{https://github.com/zhuang-li/few-shot-semantic-parsing}}. It implements a designated transition system to generate templates, followed by filling the slot variables with values extracted from utterances. To address the challenges in the few-shot setting, we proposed three training methods, detailed in Sec. \ref{sec:training}.

Many LFs differ only in mentioned atoms, such as entities and attribute values. An LF template is created by replacing the atoms in LFs with typed slot variables. As an example, the LF template of our example in Sec. \ref{sec:intro} is created by substituting i) a typed atom variable $v_e$ for the entity ``atlanta:ci''; ii) a shared variable name $v_a$ for all variables ``$\$0$`` and ``$\$1$``.

{\small
\begin{center}
    \begin{tabular}{c r}
    \label{example_template}
    \emph{(lambda $v_a$ e (exists $v_a$ (and ( ground\_transport $v_a$ )  }\\ 
     \emph{(to\_city $v_a$ $v_e$ )(from\_airport $v_a$ $v_e$) ( =(ground\_fare $v_a$ ) $v_a$ ))))} 
    \end{tabular}
\end{center}
}



\begin{table}[t]
{\small
  \begin{center}
\begin{tabular}{l l} 
\hline\hline 
\textbf{t} & \textbf{Actions} \\ 
\hline 
$t_1$ & \gen{(ground\_transport $v_a$)}  \\
$t_2$ &  \gen{(to\_city $v_a$ $v_e$)}  \\
$t_3$ &  \gen{(from\_airport $v_a$ $v_e$)}\\
$t_4$ &  \gen{(= (ground\_fare $v_a$) $v_a$)}  \\
$t_5$ &  \reduce{and :- NT NT NT NT}  \\
$t_6$ &  \reduce{exists :- $v_a$ NT}\\
$t_7$ &  \reduce{lambda :- $v_a$ e NT}\\
\hline 
\end{tabular}
\caption{An example action sequence.}
\label{fig:tree_example}
  \end{center}
 }
\end{table}

\noindent Formally, let $\vx = \{x_1, ... , x_n\}$ denote an NL utterance, and its LF is represented as a semantic tree $\vy = (\mathcal{V}, \mathcal{E})$, where $\mathcal{V} = \{v_1, ... , v_m\}$ denotes the node set with $v_i \in \mathcal{V}$, and $\mathcal{E} \subseteq \mathcal{V} \times \mathcal{V} $ is its edge set. 
The node set $\mathcal{V} = \mathcal{V}_p \cup \mathcal{V}_v$ is further divided into a template predicate set $\mathcal{V}_p$ and a slot value set $\mathcal{V}_v$. A template predicate node represents a predicate symbol or a term, while a slot value node represents an atom mentioned in utterances. Thus, a semantic tree $\vy$ is composed of an abstract tree $\tau_{\vy}$ representing a template and a set of slot value nodes $\mathcal{V}_{v,\vy}$ attaching to the abstract tree.

In the few-shot setting, we are provided with a train set $\mathcal{D}_{train}$, a support set $\mathcal{D}_s$, and a test set $\mathcal{D}_{test}$. Each example in either of those sets is an utterance-LF pair $(\vx_i, \vy_i)$. The new predicates appear only in $\mathcal{D}_s$ and $\mathcal{D}_{test}$ but \textit{not} in $\mathcal{D}_{train}$. For $K$-shot learning, there are $K$ $(\vx_i, \vy_i)$ per each new predicate $p$ in $\mathcal{D}_s$. Each new predicate appears also in the test set. The goal is to maximize the accuracy of estimating LFs given utterances in $\mathcal{D}_{test}$ by using a parser trained on $\mathcal{D}_{train} \cup \mathcal{D}_{s}$. 

\subsection{Transition System}
We apply the transition system~\cite{cheng2019executableParser} to perform a sequence of transition actions to generate the template of a semantic tree. The transition system maintains partially-constructed outputs using a \textit{stack}. The parser starts with an empty stack. At each step, it performs one of the following transition actions to update the parsing state and generate a tree node. The process repeats until the stack contains a complete tree. 

\begin{itemize}
    \item \textbf{\gen{$y$}} creates a new leaf node $y$ and pushes it on top of the stack.
    \item \textbf{\reduce{$r$}}. The reduce action identifies an implication rule \textit{ $\text{head} :- \text{body}$}. The rule body is first popped from the stack. A new subtree is formed by attaching the rule head as a new parent node to the rule body . Then the whole subtree is pushed back to the stack. 
\end{itemize}
Table \ref{fig:tree_example} shows such an action sequence for generating the above LF template. Each action produces \textit{known} or \textit{new} predicates. 

\subsection{Base Parser}
\label{sec:nn_model}


\parser generates an LF in two steps: i) template generation, ii) slot filling. The base architecture largely resembles~\cite{cheng2019executableParser}.

\paragraph{Template Generation} Given an utterance, the task is to generate a sequence of actions $\mathbf{a} = a_1, ..., a_k$ to build an abstract tree $\tau_{\vy}$.

We found out LFs often contain idioms, which are frequent subtrees shared across LF templates. Thus we apply a \textit{template normalization} procedure in a similar manner as~\cite{iyer2019learningIdiomsSP} to preprocess all LF templates. It collapses idioms into single units such that all LF templates are converted into a compact form.

The neural transition system consists of an encoder and a decoder for estimating action probabilities.
\begin{small}
\begin{equation}
    P(\mathbf{a} | \vx ) = \prod_{t = 1}^{|\mathbf{a}|} P(a_t | \va_{<t},\vx)
\end{equation}
\end{small}

\textit{Encoder} We apply a bidirectional Long Short-term Memory (LSTM) network~\cite{gers1999LSTM} to map a sequence of $n$ words into a sequence of contextual word representations $\{\mathbf{e}\}_{i=1}^n$. 

\textit{Template Decoder} The decoder applies a stack-LSTM~\cite{dyer2015transition} to generate action sequences.
A stack-LSTM is an unidirectional LSTM augmented with a pointer. The pointer points to a particular hidden state of the LSTM, which represents a particular state of the stack. It moves to a different hidden state to indicate a different state of the stack.

At time $t$, the stack-LSTM produces a hidden state $\mathbf{h}^d_{t}$ by $\mathbf{h}^d_t = \text{LSTM}(\mu_{t}, \mathbf{h}^d_{t-1})$, where $\mu_{t}$ is a concatenation of the embedding of the action $\mathbf{c}_{a_{t-1}}$ estimated at time $t-1$ and the representation $\mathbf{h}_{y_{t-1}}$ of the partial tree generated by history actions at time $t-1$. 

As a common practice, $\mathbf{h}^d_{t}$ is concatenated with an attended representation $\mathbf{h}_t^a$ over encoder hidden states to yield $\mathbf{h}_t$, with
$
    \mathbf{h}_t = \mathbf{W} \begin{bmatrix}
    \mathbf{h}_t^d\\
    \mathbf{h}_t^a
    \end{bmatrix}
$, where $\mathbf{W}$ is a weight matrix and $\mathbf{h}^a_t$ is created by soft attention,
\begin{equation}
\label{eq:attented_rep}
    \mathbf{h}^a_t = \sum_{i=1}^n P(\mathbf{e}_i| \mathbf{h}^d_t)\mathbf{e}_i
\end{equation}
We apply dot product to compute the normalized attention scores $P(\mathbf{e}_i| \mathbf{h}^d_t)$ 
~\cite{luong2015effective}. The \textit{supervised attention}~\cite{rabinovich2017semanticParsingSupervisedAttention,yin2018tranx} is also applied to facilitate the learning of attention weights. Given $\mathbf{h}_t$, the probability of an action is estimated by:
\begin{equation}
\label{eq:action_prob}
P(a_t | \mathbf{h}_t) = \frac{\exp(\mathbf{c}_{a_t}^{\intercal} \mathbf{h}_t)}{\sum_{a' \in \mathcal{A}_t}\exp(\mathbf{c}_{a'}^{\intercal} \mathbf{h}_t)}
\end{equation}
where $\mathbf{c}_{a}$ denotes the embedding of action $a$, and $\mathcal{A}_t$ denotes the set of applicable actions at time $t$. 
The initialization of those embeddings will be explained in the following section.

\paragraph{Slot Filling}
A tree node in a semantic tree may contain more than one slot variables due to template normalization. Since there are two types of slot variables, given a tree node with slot variables, we employ a 
LSTM-based decoder with the same architecture as the \textit{Template} decoder to fill each type of slot variables, respectively. The output of such a decoder is a value sequence of the same length as the number of slot variables of that type in the given tree node. 

\section{Few-Shot Model Training}
\label{sec:training}
The few-shot setting differs from the supervised setting by having a support set in testing in addition to train/test sets. The support set contains $k$ utterance-LF pairs per new predicate, while the training set contains only known predicates. To evaluate model performance on new predicates, the test set contains LFs with both known and new predicates. Given the support set, we can tell if a predicate is known or new by checking if it only exists in the train set. 

We take two steps to train our model: i) pre-training on the training set, ii) fine-tuning on the support set. Its predictive performance is measured on the test set. 
We take the two-steps approach because i) our experiments show that this approach performs better than training on the union of the train set and the support set; ii) for any new support sets, it is computationally more time efficient than training from scratch on the union of the train set and the support set.


There is a distribution discrepancy between the train set and the support set due to new predicates, the meta-learning algorithms~\cite{snell2017prototypical,finn2017maml} suggest to simulate the testing scenario in pre-training by splitting each batch into a meta-support set and a meta-test set. The models utilize the information (e.g. prototype vectors) acquired from the meta-support set to minimize errors on the meta-test set. In this way, the meta-support and meta-test sets simulate the support and test sets sharing new predicates. 

However, we cannot directly apply such a training procedure due to the following two reasons. First, each LF in the support and test sets is a mixture of both known predicates and new predicates. To simulate the support and test sets, the meta-support and meta-test sets should include both types of predicates as well. We cannot assume that there are only one type of predicates.
Second, our preliminary experiments show that if there is sufficient training data, it is better off training action embeddings of known predicates $\mathbf{c}$ (Eq. \eqref{eq:action_prob}) in a supervised way, while action embeddings initialized by a metric-based meta-learning algorithm~\cite{snell2017prototypical} perform better for rarely occurred new predicates. Therefore, we cope with the differences between known and new predicates by using a customized initialization method in fine-tuning and a designated pre-training procedure to mimic fine-tuning on the train set. In the following, we introduce fine-tuning first because it helps understand our pre-training procedure.


\subsection{Fine-tuning}
During fine-tuning, the model parameters and the action embeddings in Eq. \eqref{eq:action_prob} for known predicates are obtained from the pre-trained model. The embedding of actions that produce new predicates  $\mathbf{c}_{a_t}$ are initialized using prototype vectors as in prototypical networks~\cite{snell2017prototypical}.
The prototype representations act as a type of regularization, which shares the similar idea as the deep learning techniques using pre-trained models.

A prototype vector of an action $a_t$ is constructed by using the hidden states of the \textit{template} decoder collected at the time of predicting $a_t$ on a support set. Following~\cite{snell2017prototypical}, a prototype vector is built by taking the mean of such a set of hidden states $\mathbf{h}_t$.
\begin{equation}
    \mathbf{c}_{a_t} = \frac{1}{|M|}\sum_{\mathbf{h}_t \in M} \mathbf{h}_t
    \label{eq:action_embed_proto}
\end{equation}
where $M$ denotes the set of all hidden states at the time of applying the action $a_t$. After initialization, the whole model parameters and the action embeddings are further improved by fine-tuning the model on the support set with a supervised training objective $\mathcal{L}_f$.
\begin{equation}
    \mathcal{L}_f = \mathcal{L}_s + \lambda \Omega
\end{equation}
where $\mathcal{L}_s$ is the cross-entropy loss and $\Omega$ is an attention regularization term explained below. The degree of regularization is adjusted by $\lambda \in \mathbb{R}^+$.

\paragraph{Attention Regularization} We address the poorly learned attention scores $P(\mathbf{e}_i| \mathbf{h}^d_t)$ of infrequent actions by introducing a novel attention regularization. We observe that the probability $P(a_j | x_i) = \frac{\text{count}(a_j, x_i)}{\text{count}(x_i)}$ and the character similarity between the predicates generated by action $a_j$ and the token $x_i$ are often strong indicators of their alignment. The indicators can be further strengthened by manually annotating the predicates with their corresponding natural language tokens. In our work, we adopt $1 - dist(a_j, x_i)$ as the character similarity, where $dist(a_j, x_j)$ is normalized Levenshtein distance~\cite{levenshtein1966stringDistance}. Both measures are in the range $[0,1]$, thus we apply $g(a_j, x_i) = \sigma(\cdot)P(a_j | x_i) + (1 - \sigma(\cdot)char\_sim(a_j, x_i)$ to compute alignment scores, where the sigmoid function $\sigma(\mathbf{w}^\intercal_p \mathbf{h}^d_t)$ combines two constant measures into a single score. The corresponding normalized attention scores is given by
\begin{equation}
    P'(x_i | a_k) = \frac{g(a_k, x_i)}{\sum_{j=1}^n g(a_k, x_j)}
\end{equation}
The attention scores $P(x_i | a_k)$ should be similar to $P'(x_i | a_k)$. Thus, we define the regularization term as $\Omega = \sum_{ij} | P(x_i | a_j) - P'(x_i | a_j)|$ during training.

\subsection{Pre-training}
The pre-training objective are two-folds: i) learn action embeddings for known predicates in a supervised way, ii) ensure our model can quickly adapt to the actions of new predicates, whose embeddings are initialized by prototype vectors before fine-tuning.
 
\paragraph{Predicate-dropout} Starting with randomly initialized model parameters, we alternately use one batch for the meta-loss $\mathcal{L}_m$ and one batch for optimizing the supervised loss $\mathcal{L}_s$. 

In a batch for $\mathcal{L}_m$, we split the data into a meta-support set and a meta-test set. In order to simulate existence of new predicates, we randomly select a subset of predicates as "new", thus their action embeddings $\mathbf{c}$ are replaced by prototype vectors constructed by applying Eq. \eqref{eq:action_embed_proto} over the meta-support set. The actions of remaining predicates keep their embeddings learned from previous batches. The resulted action embedding matrix $\mathbf{C}$ is the combination of both.
\begin{equation}
\label{eq:action_embed_meta}
    \mathbf{C} = (1 - \mathbf{m}^\intercal) \mathbf{C}_s + \mathbf{m}^\intercal \mathbf{C}_m
\end{equation}
where $\mathbf{C}_s $ is the embedding matrix learned in a supervised way, and $\mathbf{C}_m$ is constructed by using prototype vectors on the meta-support set. The mask vector $\mathbf{m}$ is generated by setting the indices of actions of the "new" predicates to ones and the other to zeros. We refer to this operation as \textit{predicate-dropout}. The training algorithm for the meta-loss is summarised in Algorithm \ref{algo:sup}.

In a batch for $\mathcal{L}_s$, we update the model parameters and all action embeddings with a cross-entropy loss $\mathcal{L}_s$, together with the attention regularization. Thus, the overall training objective becomes
\begin{equation}
   \mathcal{L}_p = \mathcal{L}_m + \mathcal{L}_s + \lambda \Omega
\end{equation}
\paragraph{Pre-training smoothing} Due to the new predicates, the number of candidate actions during the prediction of fine-tuning and testing is larger than the one during pre-training. That leads to distribution discrepancy between pre-training and testing. To minimize the differences, we assume a prior knowledge on the number of actions for new predicates by adding a constant $k$ to the denominator of Eq. \eqref{eq:action_prob} when estimating the action probability $P(a_t | \mathbf{h}_t)$ during pre-training. 
\begin{equation}
\label{eq:smooth}
P(a_t | \mathbf{h}_t) = \frac{\exp(\mathbf{c}_{a_t}^{\intercal} \mathbf{h}_t)}{\sum_{a' \in \mathcal{A}_t}\exp(\mathbf{c}_{a'}^{\intercal} \mathbf{h}_t) + k}
\end{equation}
We do not consider this smoothing technique during fine-tuning and testing.  
Despite its simplicity, the experimental results show a significant performance gain on benchmark datasets.
\input{alg-dropout.tex}





%% file: alg-dropout.tex
\begin{algorithm}[t]
{\small
\SetKwData{Left}{left}\SetKwData{This}{this}\SetKwData{Up}{up}
\SetKwFunction{Union}{Union}\SetKwFunction{FindCompress}{FindCompress}
\SetKwInOut{Input}{Input}\SetKwInOut{Output}{Output}
\SetAlgoLined
\Input{Training set $\mathcal{D}$, supervisely trained action embedding $\mathcal{C}_s$, number of meta-support examples $k$, number of meta-test examples $n$ per one support example, predicate-dropout ratio $r$}
\Output{The loss $\mathcal{L}_m$.}
    Extract a template set $\mathcal{T}$ from the training set $\mathcal{D}$\\
    Sample a subset $\mathcal{T}_i$ of size $k$ from $\mathcal{T}$\\ 
    $S$ := $\emptyset$ \textcolor{blue}{\# \textit{meta-support set}}\\ 
    $Q$ := $\emptyset$  \textcolor{blue}{\# \textit{meta-test set}}\\
    \For{t in  $\mathcal{T}_i$}{
    Sample a meta-support example $s'$ with template $t$ from $\mathcal{D}$ without replacement\\ 
    Sample a meta-test set $Q'$ of size $n$ with template $t$ from $\mathcal{D}$ \\
    $S = S \cup s'$\\
    $Q = Q \cup Q'$\\
    }
    Build a prototype matrix $\mathcal{C}_m$ on $S$\\
    Extract a predicate set $\mathcal{P}$ from $S$\\ 
    Sample a subset $\mathcal{P}_s$ of size $r\times |\mathcal{P}|$ from $\mathcal{P}$ as new predicates \\ 
    Build a mask $\mathbf{m}$ using $\mathcal{P}_s$\\
    With $\mathcal{C}_s$, $\mathcal{C}_m$ and $\mathbf{m}$, apply Eq. \eqref{eq:action_embed_meta} to compute $\mathbf{C}$\\
    Compute $\mathcal{L}_m$, the cross-entropy on $Q$ with $\mathbf{C}$\\

}
\caption{Predicate-Dropout
}
\label{algo:sup}
\end{algorithm}

%% file: sec4_expr.tex
\paragraph{Datasets.}
We use three semantic parsing datasets: \Jobs, \Geo, and \Atis. 
\Jobs contains 640 question-LF pairs in Prolog about job listings. \geoquery~\cite{zelle1996geoQuery} and \Atis~\cite{price1990atis} include 880 and 5,410 utterance-LF pairs in lambda calculas about US geography and flight booking, respectively. The number of predicates in \Jobs, \geoquery, \Atis is 15, 24, and 88, respectively. All atoms in the datasets are anonymized as in~\cite{dong2016language}.

For each dataset, we randomly selected $m$ predicates as the new predicates, which is 3 for \Jobs, and 5 for \Geo and \Atis. Then we split each dataset into a train set and an \textit{evaluation} set. And we removed the instances, the template of which is unique in each dataset. The number of such instances is around 100, 150 and 600 in \Jobs, \Geo, and \Atis. The ratios between the evaluation set and the train set are 1:4, 2:5, and 1:7 in \jobs, \geoquery, and \atis, respectively. Each LF in an evaluation set contains at least a new predicate, while an LF in a train set contains only known predicates. To evaluate $k$-shot learning, we build a support set by randomly sampling $k$ pairs per new predicate without replacement from an \textit{evaluation} set, and keep the remaining pairs as the test set. To avoid evaluation bias caused by randomness, we repeat the above process six times to build six different splits of support and test set from each evaluation set. One for hyperparameter tuning and the rest for evaluation. We consider at most 2-shot learning due to the limited number of instances per new predicate in each evaluation set.

\paragraph{Training Details.}
\label{para:training}
We pre-train our parser on the training sets for \{80, 100\} epochs with the Adam optimizer~\cite{kingma2014adam}. The batch size is fixed to 64. The initial learning rate is 0.0025, and the weights are decayed after 20 epochs with decay rate 0.985. The predicate dropout rate is 0.5. The smoothing term is set to \{3, 6\}. The number of meta-support examples is 30 and the number of meta-test examples per support example is 15.
The coefficient of attention regularization is set to 0.01 on \jobs and 1 on the other datasets. 
We employ the 200-dimensional GLOVE embedding~\cite{pennington2014glove} 
to initialize the word embeddings for utterances. The hidden state size of all LSTM models~\cite{hochreiter1997long} is 256. 
During fine-tuning, the batch size is 2, the learning rates and the epochs are selected from \{0.001, 0.0005\} and \{20, 30, 40, 60, 120\}, respectively. 


\input{meta-result.tex}
\paragraph{Baselines.}
We compared our methods with five competitive baselines, \seqseq with attention~\cite{luong2015effective}, 
\coarsefine~\cite{dong2018coarse2fine}, \irnet~\cite{guo2019irnet}, \ptmaml~\cite{huang2018metaLearning} and \da~\cite{li2020domain}. \coarsefine is the best performing supervised model on the standard split of \Geo and \Atis datasets. 
\ptmaml is a few-shot learning semantic parser that adopts Model-Agnostic Meta-Learning~\cite{finn2017maml}. We adapt \ptmaml in our scenario by considering a group of instances that share the same template as a pseudo-task. \da is the most recently proposed neural semantic parser applying domain adaptation techniques. 
\irnet is the strongest semantic parser that can generalize to unseen database schemas. In our case, we consider a list of predicates in support sets as the columns of a new database schema and incorporate the schema encoding module of \irnet into the encoder of our base parser. We choose \irnet over \ratsql~\cite{wang2019rat} because \irnet achieves superior performance on our datasets.


We consider three different supervised learning settings. First, we pre-train a model on a train set, followed by fine-tuning it on the corresponding support set, coined \textit{pt}. Second, a model is trained on the combination of a train set and a support set, coined \textit{cb}. Third, the support set in \textit{cb} is oversampled by 10 times and 5 times for one-shot and two-shot respectively, coined \textit{os}. 
\paragraph{Evaluation Details.}
The same as prior work~\cite{dong2018coarse2fine,li2020domain}, we report accuracy of exactly matched LFs as the main evaluation metric. 

To investigate if the results are statistically significant, we conducted the Wilcoxon signed-rank test, which assesses whether 
our model consistently performs better than another baseline across \textit{all} evaluation sets. It is considered superior than t-test in our case, because it supports comparison across different support sets and does not assume normality in data~\cite{demvsar2006statistical}. We include the corresponding $p$-values in our result tables.

\subsection{Results and Discussion}

Table \ref{table:meta} shows the average accuracies and significance test results
of all parsers compared on all three datasets. Overall, \parser outperforms all baselines with at least 2\% on average in terms of accuracy in both one-shot and two-shot settings. The results are statistically significant w.r.t. the strongest baselines, \irnet(\comb) and \coarsefine(\pt). The corresponding p-values are 0.00276 and 0.000148, respectively. Given one-shot example on \jobs,
our parser achieves 7\% higher accuracy than the best baseline, and the gap is 4\% 
on \Geo with two-shots examples. In addition, none of the SOTA baseline parsers can consistently outperform other SOTA parsers when there are few parallel data for new predicates. In one-shot setting, the best supervised baseline \irnet (\comb) can achieve the best results on \geoquery and \jobs among all baselines, and on two-shot setting, it performs best only on \geoquery.
It is also difficult to achieve good performance by adapting the existing meta-learning or transfer learning algorithms to our problem, as evident by the moderate performance of \ptmaml and \da on all datasets.

The problems of few-shot learning demonstrate the challenges imposed by infrequent predicates. There are significant proportions of infrequent predicates on the existing datasets. For example, on \geoquery, there are 10 predicates contributing to only 4\% of the total frequency of all 24 predicates, while the top two frequent predicates amount to 42\%. As a result, the SOTA parsers achieve merely less than 25\% and 44\% of accuracy with one-shot and two-shots examples, respectively. In contrast, those parsers achieve more than 84\% accuracy on the standard splits of the same datasets in the supervised setting.


Infrequent predicates in semantic parsing can also be viewed as a class imbalance problem, when support sets and train sets are combined in a certain manner. In this work, the ratio between the support set and the train set in \jobs, \geoquery, and \atis is 1:130, 1:100, and 1:1000,  respectively. Different models prefer different ways of using the train sets and support sets. The best option for \coarsefine and \seqseq is to pre-train on a train set followed by fine-tuning on the corresponding support set, while \irnet favors oversampling in two-shot setting.

\paragraph{Ablation Study}
\input{ablation-result.tex}
We examine the effect of different components of our parser by removing each of them individually and reporting the corresponding average accuracy. As shown in Table \ref{table:ablation}, removing any of the components almost always leads to statistically significant drop of performance. The corresponding p-values are all less than 0.00327.

To investigate predicate-dropout, we exclude either supervised-loss during pre-training (-sup) or initialization of new predicate embeddings by prototype vectors before fine-tuning (-proto). It is clear from Table \ref{table:ablation} that ablating either supervisely trained action embeddings or prototype vectors hurts performance severely.  

We further study the efficacy of attention regularization by removing it completely (-reg), removing only the string similarity feature (-strsim), or conditional probability feature (-cond). Removing the regularization completely degrades performance sharply except on \jobs in the one-shot setting. Our further inspection shows that model learning is easier on \jobs than on the other two datasets. Each predicate in \jobs almost always aligns to the same word across examples, while a predicate can align with different word/phrase in different examples in \geoquery and \atis. The performance drop with -strsim and -cond indicates that we cannot only reply on a single statistical measure for regularization. For instance, we cannot always find predicates take the same string form as the corresponding words in input utterances. In fact, the proportion of predicates present in input utterances is only 42\%, 38\% and 44\% on \jobs, \atis, and \geoquery, respectively.

Furthermore, without pre-training smoothing (- smooth), the accuracy drops at least 1.6\% in terms of mean accuracy on all datasets. Smoothing enables better model parameter training by more accurate modelling in pre-training.


\paragraph{Support Set Analysis}
\begin{figure}[t] 
\centering
\includegraphics[width=1\textwidth]{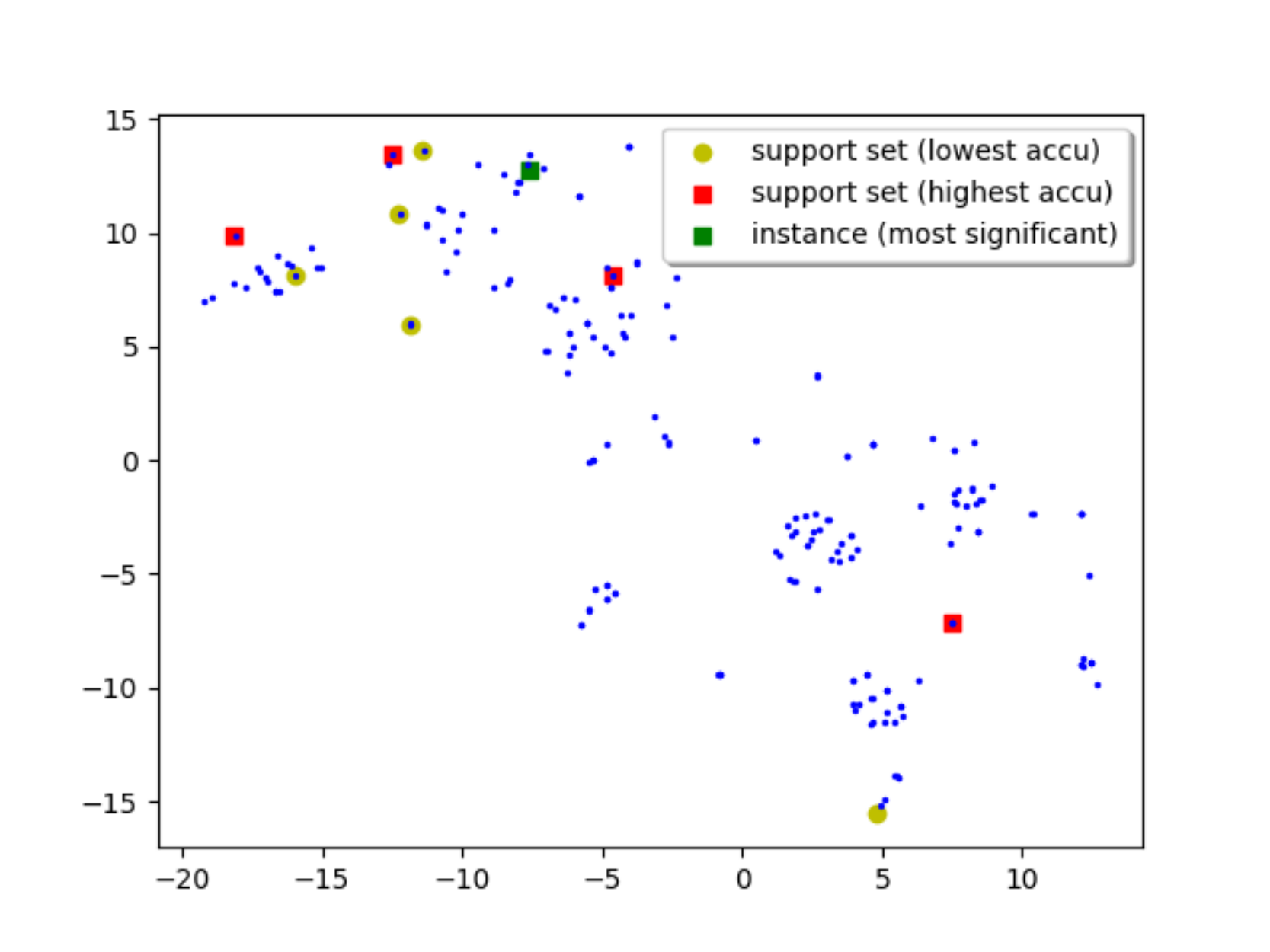} 
\caption{(Round) The support set with the lowest accuracy. (Box) The support set with the highest accuracy.}
\label{fig:utter_encode}
\vspace{-6mm}
\end{figure}
We observe that all models consistently achieve high accuracy on certain support sets of the same dataset, while obtaining low accuracies on the other ones. 
We illustrate the reasons of such effects by plotting the evaluation set of \geoquery. Each data point in Figure \ref{fig:utter_encode} depicts an representation, which is generated by the encoder of our parser after pre-training. We applied T-SNE~\cite{maaten2008visualizing} for dimension reduction. We highlight two support sets used in the one-shot setting on \geoquery.
All examples in the highest performing support set tend to scatter evenly and cover different dense regions in the feature space, while the examples in the lowest performing support set are far from a significant number of dense regions. Thus, the examples in good support sets are more representative of the underlying distribution than the ones in poor support sets. When we leave out each example in the highest performing support set and re-evaluate our parser each time, we observe that the good ones (e.g. the green box in Figure \ref{fig:utter_encode}) locate either in or close to some of the dense regions.

%% file: meta-result.tex

\begin{table*}[ht]
\small
\begin{tabular}[t]{@{} |l|ccc||ccc|l|}
\toprule
         & \Jobs & \Geo & \Atis & \Jobs & \Geo & \Atis & p-values\\ 
\midrule
\hline         
\seqseq (\pt)  & 11.27  &  20.00      &   17.23  & 14.58   &  33.01     &   18.76 & 3.32e-04 \\
\seqseq (\comb)  & 11.70   &  7.64     &   2.25  & 21.49   &  14.36      &   7.91 & 6.65e-06 \\
\seqseq (\os)  & 14.18   &  11.38      &   4.45  & 30.46   &  33.59     &   10.17 & 5.30e-05\\
\hline
\coarsefine (\pt) & 10.91   & 24.07       &  17.44  & 13.83  & 35.63       &  21.08 &  1.48e-04 \\
\coarsefine (\comb) & 9.28  & 14.50     &  0.42  & 19.61   & 28.93       &  9.25 & 2.35e-06 \\
\coarsefine (\os) & 6.73   & 10.35        &  5.26  & 16.08   & 28.55      &  17.73 & 1.13e-05  \\
\hline
\irnet (\pt) & 16.00   & 20.00     &  17.12  & 19.06  & 35.05        &  20.11 & 2.86e-05 \\
\irnet (\comb) & 19.67   & 21.90     &  5.60  & 28.22   & 44.08      &  15.73 &2.76e-03 \\
\irnet (\os) & 14.91  & 18.78      &  4.95 & 30.84   & 40.97       &  18.05 & 2.47e-04 \\
\hline
\da     & 18.91  & 9.67      &  4.29    & 21.31 & 20.88  &   17.18 & 1.13e-06 \\
\hline
\ptmaml     & 11.64  & 9.76     &  6.83    & 17.76  & 22.52        &   12.28  & 1.73e-06 \\
\hline
Ours  & \textbf{27.09}  & \textbf{27.49}        & \textbf{19.27}    & \textbf{32.5} & \textbf{48.45}        & \textbf{22.48}  &  \\
\bottomrule

\end{tabular}%
\caption{Evaluation of learning results on three datasets. (Left) The one-shot results. (Right) The two-shot results.}

\label{table:meta}
\vspace{-3mm}
\end{table*}

%% file: ablation-result.tex
\begin{table*}[ht]
\small
\begin{tabular}{|@{} |l|ccc||ccc|l|}
\toprule
         & \Jobs & \Geo & \Atis  & \Jobs & \Geo & \Atis & p-values\\ 
\midrule
Ours  &27.09  & \textbf{27.49}         & \textbf{19.27}  & \textbf{32.50} & \textbf{48.45}       & \textbf{22.48}  &  \\
\hline
\quad - sup  & 23.63 & 18.86  & 12.91 & 26.91 & 39.51  & 14.89 & 1.44e-05 \\

\quad -  proto  & 22.91 & 18.77  & 13.24  & 29.16  & 38.93   & 16.81 & 1.77e-05 \\
\hline
\quad -  reg  & \textbf{29.27 } & 18.10 & 13.66  & 31.03 & 39.61  & 18.58 & 9.60e-04 \\

\quad -  strsim  & 22.18 & 19.62  & 10.14  & 28.41  & 47.09  & 19.98 & 9.27e-04 \\

\quad -  cond  & 23.27 & 19.05  & 9.63  & 27.66  & 40.97  & 17.50 & 4.37e-05 \\
\hline

\quad - smooth  & 24.36  & 23.60       & 15.23  & 30.84   & 44.95        & 18.71  & 3.27e-03  \\
\bottomrule
\end{tabular}%
\caption{Ablation study results. (Left) The one-shot learning results. (Right) The two-shot learning results.}
\label{table:ablation}
\vspace{-2mm}
\end{table*}



%% file: sec5_conc.tex
We propose a novel few-shot learning based semantic parser, coined \parser, to cope with new predicates in LFs. To address the challenges in few-shot learning, we propose to train the parser with a pre-training procedure involving predicate-dropout, attention regularization, and pre-training smoothing. The resulted model achieves superior results over competitive baselines on three benchmark datasets.

%% file: appendix.tex
\begin{appendices}
\section{Template Normalization}
\label{app:normalize_trees}


\input{alg-tree-norm.tex}

\input{tab-example-trans.tex}

Many LF templates in the existing corpora have shared subtrees in the corresponding abstract semantic trees. The tree normalization algorithm aims to treat those subtrees as single units. The identification of such shared structured is conducted by finding frequent subtrees. Given an LF dataset, the \textit{support} of a tree $t$ is the number of LFs that it occurs as a subtree. We call a tree \textit{frequent} if its support is greater and equal to a pre-specified minimal support.

We also observe that in an LF dataset, some frequent subtrees always have the same supertree. For example, \pre{ground\_fare \$1} is always the child of \textit{=( $\dots$, \$0 )} in the whole dataset. We call a subtree \textit{complete} w.r.t. a dataset if any of its supertrees in the dataset occur significantly more often than that subtree. Another observation is that some tree nodes have fixed siblings. In order to check if two tree nodes sharing the same root are fixed siblings, we merge the two tree paths together. If the merged tree has the same support as that of the of the two trees, we call the two trees pass the fixed sibling test. In the same manner, we collapse tree nodes with fixed siblings, as well as their parent node into a single tree node to save unnecessary parse actions.

Thus, the normalization is conducted by collapsing a frequent complete abstract subtree into a tree node. We call a tree \textit{normalized} if all its frequent complete abstract subtrees are collapsed into the corresponding tree nodes. The pseudocode of the tree normalization algorithm is provided in Algorithm~\ref{algo:normalization}.

\section{One Example Transition Sequence}
As in Table \ref{table:example_transition}, we provide an example transition sequence to display the stack states and the corresponding action sequence when parsing the utterance in Introduction "how much is the ground transportation between atlanta and downtown?”.

\end{appendices}

%% file: alg-tree-norm.tex
\begin{algorithm}[t]
{\small
\SetKwData{Left}{left}\SetKwData{This}{this}\SetKwData{Up}{up}
\SetKwFunction{Union}{Union}\SetKwFunction{FindCompress}{FindCompress}
\SetKwInOut{Input}{Input}\SetKwInOut{Output}{Output}
\SetAlgoLined
\Input{A set of abstract trees $\mathcal{T}$, a minimal support $\tau$}
\Output{A set of normalized trees}
$O$ := mapping of subtrees to their occurrences in $\mathcal{T}$. \\
\For{tree $t$ in $\mathcal{T}$}{
    update occurrence of all leaf nodes $v$ of $t$ to $O[v]$
}
\While{$O$ updated with new trees}{
\For{tree $t$, occur\_list $l$ in $\mathcal{O}$}{
   build occurrence list $l'$ for supertree $t'$ of $t$\\
   \If{size($l'$) $\geq$ size($l$)}
   {
     $O[t'] = l'$ 
   }
}
}
\For{tree $t$, occur\_list $l$ in $\mathcal{O}$}{
    \If{size($l$) $\geq \tau$}
    {
        collapse $t$ into a node for all $t'$ in $l$.
    }
}
}
\caption{Template Normalization}
\label{algo:normalization}
\end{algorithm}

%% file: tab-example-trans.tex
\begin{table*}[!b] 
\centering 
{\small
\begin{tabular}{l c l} 
\hline\hline 
\textbf{t} & \textbf{Stack}& \textbf{Action} \\ 
\hline 
$t_1$ & [] & \gen{(ground\_transport $v_a$)}  \\
$t_2$ & [(ground\_transport $v_a$)] & \gen{(to\_city $v_a$ $v_e$)}  \\
$t_3$ & [(ground\_transport $v_a$), (to\_city $v_a$ $v_e$)] & \gen{(from\_airport $v_a$ $v_e$)}\\
$t_4$ & [(ground\_transport $v_a$), (to\_city $v_a$ $v_e$), (from\_airport $v_a$ $v_e$)] & \gen{(= (ground\_fare $v_a$) $v_a$)}  \\
$t_5$ & [(ground\_transport $v_a$), (to\_city $v_a$ $v_e$),&\\ & (from\_airport $v_a$ $v_e$), (= (ground\_fare $v_a$) $v_a$)] & \reduce{and :- NT NT NT NT}  \\
$t_6$ & [(and (ground\_transport $v_a$) (to\_city $v_a$ $v_e$) &\\ &(from\_airport $v_a$ $v_e$) (= (ground\_fare $v_a$) $v_a$))] & \reduce{exists :- $v_a$ NT}\\
$t_7$ & [(exists $v_a$ (and (ground\_transport $v_a$) (to\_city $v_a$ $v_e$) &\\ &(from\_airport $v_a$ $v_e$) (= (ground\_fare $v_a$) $v_a$)))] &\reduce{lambda :- $v_a$ e NT}  \\

$t_8$ & [(lambda $v_a$ e (exists $v_a$ (and (ground\_transport $v_a$) (to\_city $v_a$ $v_e$) &\\ &(from\_airport $v_a$ $v_e$) (= (ground\_fare $v_a$) $v_a$))))] & \\  [1ex] 
\hline 
\end{tabular}
}
\caption{The transition sequence for LF template parsing "how much is the ground transportation between atlanta and downtown?".}
\label{table:example_transition} 
\end{table*}
\FloatBarrier